\documentclass[10pt,twocolumn,letterpaper]{article}

\usepackage[pagenumbers]{cvpr} 

\usepackage{graphicx}
\usepackage{amsmath}
\usepackage{amssymb}
\usepackage{booktabs}

\usepackage{multirow}
\usepackage{pifont}
\usepackage{makecell}
\usepackage{verbatim}
\usepackage{color}
\newcommand{\cmark}{\ding{51}}%
\newcommand{\xmark}{\ding{55}}%
\usepackage[flushleft]{threeparttable}

\usepackage[pagebackref,breaklinks,colorlinks]{hyperref}

\usepackage[capitalize]{cleveref}
\crefname{section}{Sec.}{Secs.}
\Crefname{section}{Section}{Sections}
\Crefname{table}{Table}{Tables}
\crefname{table}{Tab.}{Tabs.}

\def\cvprPaperID{3916} 
\def\confName{CVPR}
\def\confYear{2023}

\begin{document}

\title{LargeKernel3D: Scaling up Kernels in 3D Sparse CNNs}

\author{Yukang Chen$^{1}$\thanks{Equal Contribution.},
~~~
Jianhui Liu$^{2*}$,~~~
Xiangyu Zhang$^{3}$,~~~
Xiaojuan Qi$^{2}$,~~~
Jiaya Jia$^{1}$
\\[0.2cm]
$^1$The Chinese University of Hong Kong~~
$^2$The University of Hong Kong~~
$^3$MEGVII Technology~~
}
\maketitle

\begin{abstract}
  Recent advance in 2D CNNs has revealed that large kernels are important. 
  However, when directly applying large convolutional kernels in 3D CNNs, severe difficulties are met, where those successful module designs in 2D become surprisingly ineffective on 3D networks, including the popular depth-wise convolution. To address this vital challenge, we instead propose the {spatial-wise partition convolution} and its large-kernel module. As a result, it avoids the optimization and efficiency issues of naive 3D large kernels.
  Our large-kernel 3D CNN network, LargeKernel3D, yields notable improvement in 3D tasks of semantic segmentation and object detection. It achieves \textbf{73.9\%} mIoU on the ScanNetv2 semantic segmentation and  {\textbf{72.8\%} NDS nuScenes object detection benchmarks, ranking \textbf{1$^{st}$} on the nuScenes LIDAR leaderboard. The performance further boosts to \textbf{74.2\%} NDS with a simple multi-modal fusion. In addition, LargeKernel3D can be scaled to 17$\times$17$\times$17 kernel size on Waymo 3D object detection.}
  For the first time, we show that large kernels are feasible and essential for 3D visual tasks. Our code and models is available at \href{https://github.com/dvlab-research/LargeKernel3D}{github.com/dvlab-research/LargeKernel3D}.
  
\end{abstract}

\section{Introduction}
\label{sec:intro}

3D Sparse convolutional neural networks~(CNNs) have been widely used as feature extractors in 3D tasks, {\em e.g.}, semantic segmentation~\cite{minkowskinet, sparseconvnet} and object detection~\cite{pvrcnn, voxel-rcnn, centerpoint}. The advantages of efficiency and convenient usage ensure its important role in various applications, such as autonomous driving and robotics.
However, 3D sparse CNNs are recently challenged by transformer-based methods~\cite{voxeltransformer,point-transformer, fast-point-transformer}, mainly from the aspect of building effective receptive fields. Both global and local~\cite{single-stride,voxeltransformer} self-attention mechanisms are able to capture context information from a large spatial scope. 
2D Vision Transformers (ViTs) also emphasize their advantages in modeling long-range dependencies~\cite{vit,swin-transformer,vision-transformer-like-cnn}. In contrast, common 3D sparse CNNs are limited in this regard. It is because the receptive fields of default 3D sparse CNN are constrained by small kernel sizes and spatial disconnection of sparse features (due to the property of submanifold sparse convolution~\cite{submanifold-sparse-conv-v2}).
\begin{figure*}[t]
\begin{center}
   \includegraphics[width=\linewidth]{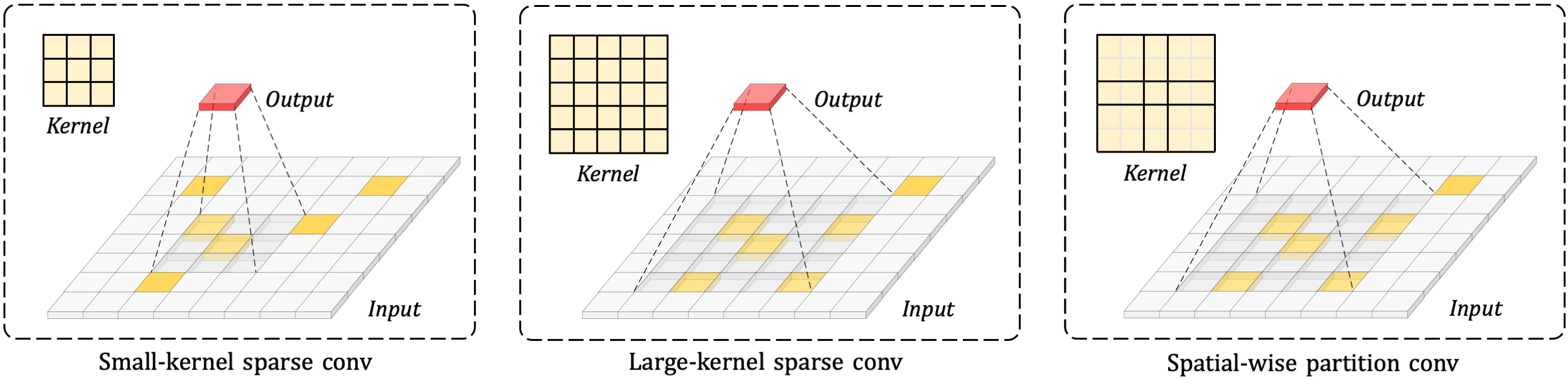}
   \caption{Sparse convolutions with different kernels. {\em Small-kernel sparse convolution} gathers features in a local area. It is efficient but discards sufficient information flow due to feature disconnection and the small scope. {\em Large-kernel sparse convolution} is capable of capturing long-range information, at the price of a large number of parameters and computation. Our proposed {\em   {spatial-wise partition convolution}} uses large kernel sizes, and shares weights among local neighbors for efficiency. We show 2D features for the simplicity sake.}
   \label{fig:large-small-kernels}
\end{center}
\end{figure*}

Literature about 2D CNNs~\cite{convnext, large-kernel,patches-all-needed} presents a series of methods, combined with large kernels, to enlarge the receptive fields and model capacity.
ConvNeXt~\cite{convnext} employs 7$\times$7 depth-wise convolution as a strong design, combing with other training techniques to challenge its Swin Transformer counterpart~\cite{swin-transformer}. RepLKNet~\cite{large-kernel} pursues extremely large kernel sizes of 31$\times$31 to boost the performance of different tasks. To ensure the effectiveness of RepLKNet~\cite{large-kernel}, additional factors, including depth-wise convolution, are also required. Other work~\cite{connection-attention-dynamic-depthwise} also emphasizes the importance of depth-wise convolution. Due to differences between 3D and 2D tasks, these methods, however, are found not a good solution for 3D sparse CNNs. 

We first analyze the difficulties of 3D large-kernel CNN design in two aspects.
The first challenge is {\em efficiency}. It is easy to understand that 3D convolution is with the cubic kernel size and computation increases fast. For example, the model size increases 10+ times when kernels change from 3$\times$3$\times$3 to 7$\times$7$\times$7.
The second difficulty exists in the {\em optimization} procedure. 3D datasets may contain only thousands of scenes, which cannot match 2D image benchmarks~\cite{imagenet, coco} in terms of scales. In addition, 3D point clouds or voxels are sparse, instead of dense images. Thus, it might be insufficient to optimize the proliferated parameters of large kernels and leads to over-fitting. 

In this paper, we propose {spatial-wise partition convolution} as the 3D large-kernel design. It is a new family of group convolution by sharing weights among spatially adjacent locations, rather than {depth-wise convolution}~\cite{mobilenet} of channel-level groups. As shown in Fig.~\ref{fig:large-small-kernels}, spatial-wise partition convolution remaps a large kernel ({\em e.g.}, 7$\times$7) as a small one ({\em e.g.}, 3$\times$3) via grouping spatial neighbors, while the absolutely large spatial size remains unchanged. With regard to the {\em efficiency} issue, it occupies few model sizes to keep parameters the same as those of small kernels. Moreover, it takes less latency, compared with plain large kernel counterparts. As for the {\em optimization} challenge, weight-sharing among spatial dimensions gives parameters more chance to update and overcome the over-fitting issue. 

To increase the detail-capturing ability of large kernels~\cite{large-kernel}, we introduce position embeddings for spatial-wise group convolution. It makes notable effects for large kernel sizes.
We name the proposed block as spatial-wise large-kernel convolution~({\em SW-LK Conv}). We compare the efficiency between plain 3D submanifold sparse convolution and ours, as shown in Tab.~\ref{tab:efficiency-comparison}. Both parameters and latency of the baseline increases dramatically, as its kernel size becomes larger, while ours is far more efficient.

SW-LK Conv can readily replace plain convolution layers in existing 3D convolutional networks. We establish large-kernel backbone networks LargeKernel3D on existing 3D semantic segmentation~\cite{minkowskinet} and object detection~\cite{voxel-rcnn,centerpoint} networks. It achieves notable improvement upon state-of-the-art methods~\cite{voxel-rcnn, centerpoint, minkowskinet}, with a small model complexity overhead. Extensive experiments validate our effectiveness on large-scale benchmarks, including ScanNetv2~\cite{scannet}, nuScenes~\cite{nuscenes}, and Waymo~\cite{waymo}. For object detection, LargeKernel3D achieves \textbf{72.8\%} NDS on nuScenes, ranking \textbf{1$^{st}$} on the nuScenes LIDAR leaderboard. Without bells and whistles, it further improves to \textbf{74.2\%} NDS in a simple voxel-wise multi-modal fusion manner.
 {More importantly, it is scalable to 17$\times$17$\times$17 kernel sizes on the large-scale Waymo 3D object detection.}

\begin{figure*}[t]
\begin{center}
   \includegraphics[width=\linewidth]{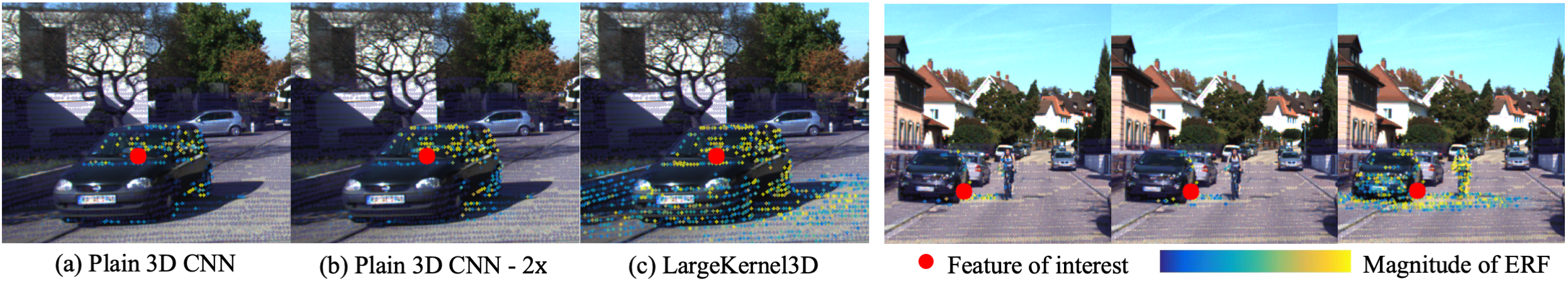}
   \caption{{\em Effective Receptive Fields}~(ERF) of plain 3D CNN, plain 3D CNN - 2$\times$, and our LargeKernel3D for 3D object detection. The plain 3D CNN backbone has insufficient ERF size, which hardly covers a nearby car. Plain 3D CNN - 2$\times$ is a deeper version with its layers doubled in each stage. More layers help little in ERF, as the disconnection between sparse features. Our LargeKernel3D obtains large ERF at both the object centers (left) and edges (right), capturing context information. More illustrations are provided in the {\em appendix}. The figure is best viewed in color and by zoom-in.}
   \label{fig:effective-receptive-fields}
\end{center}
\end{figure*}

We visualize the {\em Effective Receptive Fields}~(ERFs) of plain 3D CNNs and our LargeKernel3D in Fig.~\ref{fig:effective-receptive-fields}. It shows that deep small-kernel networks are also constrained by limited ERFs, since sparse features are spatially disconnected. Note that our large-kernel networks elegantly resolve this issue.
For the first time, we show that large-kernel CNN designs become effective on essential 3D visual tasks.

\begin{table*}[t]
\begin{center}
\caption{Efficiency comparison between plain 3D sparse convolution and our SW-LK Conv. The baseline is a submanifold sparse convolutional layer with input and output channels 16. The input data is a sparse tensor with 80,000 voxels randomly scattered and batch size 1. We test all latencies on a single 2080Ti GPU by an average of 10 times running, after a warm-up start of additional 10 times.}
\begin{tabular}{|cc|cccccccc|}
\hline
\multicolumn{2}{|c|}{Kernel Size}                   & 3$\times$3$\times$3      & 5$\times$5$\times$5      & 7$\times$7$\times$7      & 9$\times$9$\times$9       & 11$\times$11$\times$11      & 13$\times$13$\times$13      & 15$\times$15$\times$15      & 17$\times$17$\times$17       \\ \hline
\multicolumn{1}{|c|}{\multirow{2}{*}{Plain}} & Params  & 6.9 K  & 32.0 K & 87.8 K & 186.6 K & 340.7 K & 562.4 K & 864.0 K & 1.3 M    \\
\multicolumn{1}{|c|}{}                       & Latency & 2.5 ms & 4.2 ms & 8.9 ms & 17.5 ms & 31.1 ms & 55.1 ms & 81.1 ms & 106.3 ms \\ \hline
\multicolumn{1}{|c|}{\multirow{2}{*}{Ours}}  & Params  & -      & 8.9 K  & 12.4 K & 18.6 K  & 28.2 K  & 42.1 K  & 60.9 K  & 85.5 K   \\
\multicolumn{1}{|c|}{}                       & Latency & -      & 3.4 ms & 3.9 ms & 4.8 ms  & 6.2 ms  & 8.4 ms  & 11.4 ms & 15.8 ms  \\ \hline
\end{tabular}
\label{tab:efficiency-comparison}
\end{center}
\end{table*}

\section{Related Work}
\label{sec:related_work}

\vspace{0.5em}
\noindent
\textbf{Convolutional Large-kernel Networks}
Large-kernel setting has been commonly used in 2D convolutional networks~\cite{large-kernel, convnext, image-deconvolution}. ConvNeXt~\cite{convnext} combines various training techniques and large kernel sizes, to compete with Swin Transformer~\cite{swin-transformer}. RepLKNet~\cite{large-kernel} is a specific work that focuses on large-kernel designs. It reveals that the large kernel sizes are more beneficial to tasks of object detection and semantic segmentation than image classification. GCNs~\cite{large-kernel-seg} also show improvement from large-kernel designs on semantic segmentation. There are methods approximating large kernel sizes using implicit techniques, including Fourier domain transformation~\cite{global-filternet}, and continuous functions~\cite{ckconv}.  {Other methods change kernel shapes to enlarge receptive fields using dilated convolutions~\cite{deeplabv3,atrous-conv}, deformable convolutions~\cite{deformableconv, deformableconvv2} and active convolutions~\cite{active-conv}. These schemes sparsify kernels, and is hard to capture the original sparse features in 3D deep learning.}

\vspace{0.5em}
\noindent
\textbf{3D Feature Extractors}
A key challenge in 3D tasks~\cite{jiang2021guided, li2021simultaneous} is to learn effective representation from the sparse and non-uniformed 3D geometric data, {\em e.g.}, point clouds. In general, there are two kinds of 3D feature extractors. The first is to learn on point cloud directly, using a series of PointNet networks~\cite{pointnet,pointnet++}. The sampling and grouping operations in PointNet++~\cite{pointnet++} are also time-consuming. Follow-ups~\cite{point-transformer,fast-point-transformer} use grid-sampling or scene-partition to save computation in each forward and combine results from multiple runs. The second is to process point clouds with voxelization and apply 3D sparse CNNs. Due to its efficiency advantages, this stream of methods has been widely used in various 3D tasks, such as 3D segmentation~\cite{minkowskinet,chu2021icm,chu2022twist,lai2023spherical} and backbone networks in 3D object detectors~\cite{pvrcnn, voxel-rcnn,m3detr,infofocus}. 

The spatial group convolution (SGC)~\cite{sgc} is relevant to the proposed spatial-wise partition convolution. SGC is originally designed for the task of 3D scene completion. They both conduct the partition operation spatially. The difference is also essential and clear. Our method partition kernel weighs while SGC splits input features into groups. In terms of targets, SGC focuses more on efficiency improvement by changing feature sparsity in each spatial group, while our method is to facilitate large kernels in 3D tasks. Further analysis and comparisons are given in experiments.

\vspace{0.5em}
\noindent
\textbf{Vision Transformer}
Vision transformers that conduct attention computation in local areas or windows share a similar spirit with large-kernel models~\cite{large-kernel}. Swin Transformers~\cite{swin-transformer} capture features in shifted windows with sizes from 7 to 12. Its variants~\cite{cswin-transformer, swin-transformer-v2} show that larger window sizes are beneficial to performance. Focal Transformer~\cite{focal-transformer} captures fine-grained local attention with adaptive patch sizes. SST~\cite{single-stride} and Stratified Transformer~\cite{stratified-transformer} apply window-based self-attention on 3D detection and segmentation.

\vspace{0.5em}
\noindent
\textbf{Position Encoding}
Since the self-attention in transformers is permutation-equivalent, position encoding is well-designed for imbuing the network with the position information. It can be grouped into two streams of absolute and relative position encoding. For the first group, positions are encoded in a precise manner. Specifically, encoding~\cite{vaswani2017attention} is generated with the sinusoidal functions of different frequencies. They are then added to the input. In contrast, relative position encoding~\cite{shaw2018self} takes the relative relationship into consideration. It can naturally handle longer sequences and is more scalable. Most work~\cite{bello2019attention, yang2019xlnet, raffel2020exploring, wu2021rethinking} computes the relative distance between tokens and applies implicit encoding using learnable transformations. Besides, the literature~\cite{chu2021conditional} shows that convolution with padding can provide effective relative position information for relative position encoding. In addition to transformers, the recent work~\cite{fast-part-c} also shows that position embedding is beneficial to the CNNs with fourier transformation.

Considering that our spatial-wise partition convolution divides the kernel into different portions in pursuit of higher efficiency. The position information may be blurred due to weight sharing in each portion. We thus use relative position encoding as a bias to supplement the lost position information. More details are provided in Sec.~\ref{sec:method}.

\section{Revisiting 3D Sparse CNNs}
\label{sec:sparse-3d-conv}
\vspace{0.5em}
\noindent
\textbf{Preliminary}
3D sparse CNNs consist of 3D sparse convolutions, which are typically regular and submanifold sparse convolutions~\cite{submanifold-sparse-conv-v2}. We formulate 3D sparse convolutions in Eq.~\eqref{eq:spconv}.
Given a set of sparse input features $\{\mathrm{x}_{p \in P}\}$ with a number of $c_{\mathrm{in}}$ channels, the position of each feature $p\in P$ sparsely distributes in 3D space. We process these features by a convolution with kernel weights $\mathrm{w}\in \mathbb{R}^{\mathnormal{|K|}\times c_{\mathrm{in}} \times c_{\mathrm{out}}}$. For example, in the 3D coordinate space, $\mathrm{w}$ contains $c_{\mathrm{in}} \times c_{\mathrm{out}}$ spatial kernels with size 3 and $|\mathnormal{K}|=27$. The convolution process at the output position $\bar{p}$ is represented as
\begin{equation}
\label{eq:spconv}
    \mathrm{y}_{\bar{p}}=\sum_{k\in \mathnormal{K}} \mathrm{w}_k \cdot \mathrm{x}_{\bar{p}+k},
\end{equation}
where $k$ is an 3D offset distance from $\bar{p}$. $\bar{p}+k$ is the location around center $\bar{p}$. It enumerates all discrete locations in the kernel space $\mathnormal{K}$ where sparse features $\mathrm{x}_{\bar{p}+k}$ exist. 

\vspace{0.5em}
\noindent
\textbf{Convolution Types}
Regular sparse convolutions enlarge their output positions by dilating each input position $p\in P$ to the kernel shape $\mathnormal{K}$. It dramatically decreases feature density and increases the computational burden.
Thus, regular sparse convolutions with stride 2 are only employed at the first layer in each stage for down-sampling. Submanifold sparse convolutions, in contrast, fix their output positions $\bar{p}$ identical to input positions $p\in P$, which preserves the computation cost at low levels. Please refer to the original paper~\cite{submanifold-sparse-conv-v2} for more details.

Due to the efficiency advantage, submanifold sparse convolutions dominate most layers of 3D CNNs, except for the down-sampling layers. However, limited by the small local scope, it misses sufficient information flow for the spatially disconnected features. Increasing kernel sizes becomes a potential solution to relieve this issue.

\vspace{0.5em}
\noindent
\textbf{Obstacles in 3D Large Kernels}
{\em Efficiency} is the first issue in 3D large-kernel CNNs. When we increase kernel sizes, the amount of parameters and computational burden grows much faster than those of 2D CNNs. For instance, the parameter amount in one 3D convolutional layer increases from $27 \cdot \, C_{in} \cdot _{out}$ to $343 \cdot \, c_{in} \cdot c_{out}$, given its kernel size changing from 3 to 7. It is thus clear that naively enlarging 3D kernels is unreasonable.

{\em Optimization difficulty} is the second obstacle. The booming parameters requires sufficient data for learning. However, 3D datasets are commonly not in that large scale as 2D benchmarks. For example, ImageNet~\cite{imagenet} contains millions of images, while 3D datasets commonly contain only no more than one thousand scenes. 
The over-fitting issue gradually deteriorates the performance when increasing kernel sizes from 3$\times$3$\times$3 to 7$\times$7$\times$7 in MinkowskiNet-34~\cite{minkowskinet}.

{Previous 2D knowledge} is not helpful in 3D large-kernel CNNs. Recent 2D methods~\cite{large-kernel, convnext} 
show some helpful components in large-kernel CNNs, including depth-wise convolutions~\cite{mobilenet}, layer normalization~\cite{layer-norm}, GELU~\cite{gelus} activation. We examine these popular components and find that they are either ineffective or even harmful in 3D CNNs.
To demonstrate this finding, we use MinkowskiNet-34~\cite{minkowskinet} to benchmark and train on the 3D semantic segmentation dataset of ScanNetv2~\cite{scannet}.
We change the original activation function from ReLU to GELU, batch normalization to layer normalization, and plain convolution to depth-wise operation, respectively. 
Directly using these components makes the final system yield performance drop on both small and large kernel networks. It requires us to seek other ways to construct 3D large-kernel CNNs. 

\section{3D Large-kernel Convolutional Network}
\label{sec:method}

\subsection{Spatial-wise Partition Convolution}
\label{sec:designs-3D-large-kernel-CNNs}

A standard convolution kernel can be viewed as a 3D matrix, which consists of input channels $C_\mathrm{in}$, output channels $C_\mathrm{out}$, and the spatial kernel dimension $\mathnormal{|K|}$. Taking kernel size as $k$, the volume of spatial kernel dimension $\mathnormal{|K|}=k\times k$ for 2D convolutions and $k\times k\times k$ for 3D convolutions. Depth-wise convolutions share weights along channel dimensions, where group numbers equal to input channels. Point-wise convolutions fix kernel size as 1, which is common as a conjunction layer to depth-wise convolutions to adjust output channels.

Different from 2D methods, which rely on depth-wise convolutions for performance boosting~\cite{connection-attention-dynamic-depthwise, convnext} and accuracy-FLOPs trade-offs~\cite{large-kernel}, we empirically find that depth-wise convolutions are not beneficial to 3D tasks, no matter whether point-wise convolutions are included. 

We instead propose spatial-wise partition convolution for 3D large-kernel CNNs. It shares weights among spatial dimension $\mathnormal{K}$ on convolutional kernels, instead of among channel dimension. It is also different from SGC~\cite{sgc}, which partitions spatial groups on input features. Specially, in Eq.~\eqref{eq:spconv}, $w_k$ shares the same values in a local area $k \in K$. We thus group the original large-kernel convolution from  {$7\times 7$} into $3\times 3$, by sharing weights among spatial neighbors.   {Spatial-wise partition convolution} is specially designed for 3D tasks and has the following advantages in terms of efficiency and performance. 
\begin{figure}[t]
\begin{center}
   \includegraphics[width=\linewidth]{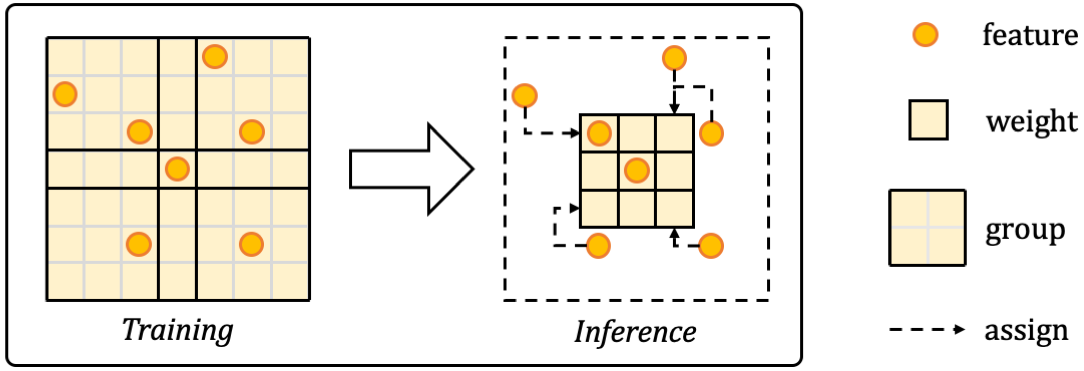}
   \caption{Illustration on {spatial-wise partition convolution}. In the plain large-kernel convolution, due to the sparsity of input features, only a small proportion of weights are involved and updated in each training step. The   {spatial-wise partition convolution} relieves this issue by sharing weights among spatial neighbors. It increases the chance for weights to be optimized. During inference, we reformulate the spatial-wise partition convolutions into small ones, with feature assignment in a large scope.}
   \label{fig:spatial-wise-conv}
\end{center}
\end{figure}
\begin{figure*}[t]
\begin{center}
   \includegraphics[width=0.9\linewidth]{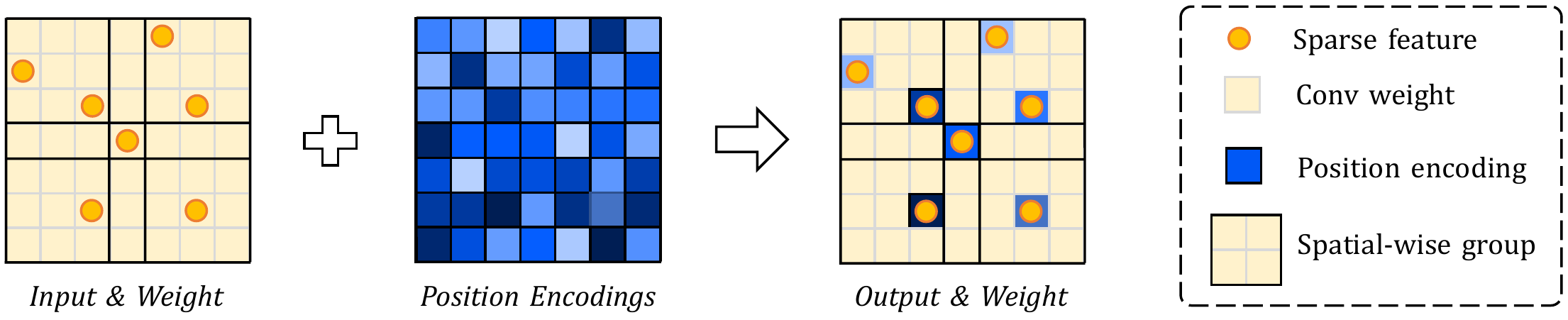}
   \caption{Structure of spatial-wise large-kernel convolution~(SW-LK Conv). It consists of a large-kernel   {spatial-wise partition convolution} and a learnable position embedding. The position embedding is used to make up the detail-capturing ability of large kernels.}
   \label{fig:training-inference}
\end{center}
\end{figure*}

Regarding {\em efficiency}, when naively enlarging the kernel size from 3$\times$3$\times$3 to 7$\times$7$\times$7 on MinkowskiNet-34~\cite{minkowskinet}, both model size and latency have a sharp increase by multiple times.
In contrast, our {spatial-wise partition convolution} avoids increasing parameters and introduces limited latency overhead. 
The advantage of our method over plain large kernels attributes to the specific implementation of sparse convolution. For a {spatial-wise partition convolution}, we directly use the small-kernel layer during inference, and enlarge its feature-assign areas to the large-kernel scope. As shown in Fig.~\ref{fig:spatial-wise-conv}, thanks to the weight-sharing operation, it much saves the multiplication, {\em e.g.}, from $343$ to $27$ times. We implement this operation upon the open-sourced {\em spconv} library and omit the details here. Tab.~\ref{tab:comparisons-convolutions-scannet} shows that our implementation on {spatial-wise partition convolution} is far more efficient than the plain large-kernel baseline. The {\em optimization} difficulty of large-kernel CNNs is relieved with the help of {spatial-wise partition convolutions}, which constrains model sizes to a limited number. It avoids requiring a large amount of data for training and the over-fitting issue.

\subsection{Kernel-wise Position Encoding}
\label{sec:position-encoding}

Considering that the spatial-wise partition convolution is designed in a sharing weights manner against the spatial sparsity. Although this design is efficient, there is still an issue: voxels within an area share the same weight, which results in local details blurred. This phenomenon is further amplified with an increase of kernel sizes. To address this problem, we propose the kernel-wise position embedding. We illustrate this design in Fig.~\ref{fig:training-inference}.

Particularly, we initialize position weights $\mathrm{e}\in \mathbb{R}^{\mathnormal{K}\times c_{\mathrm{in}}}$, which corresponds to the convolution kernel. During the convolution process, we let the input features query the position weights of the equivalent position and add them together. This process modifies Eq.~\eqref{eq:spconv} to

\begin{equation}
\label{eq:position encoding}
    \mathrm{y}_{\bar{p}}=\sum_{k\in \mathnormal{K}} \mathrm{w}_k \cdot (\mathrm{x}_{\bar{p}+k} + \mathrm{e}_{k})\,.
\end{equation}

This simple design essentially adds bias with relative position information to the input features. There is almost no extra calculation and additional parameters. It is a decent solution to the position insensitivity problem caused by sharing weights, especially for the extremely large kernels, {\em e.g.}, 17$\times$17$\times$17. For more experimental results and details, please refer to the Sec.~\ref{sec:experiments}.

\subsection{Large-kernel Architecture}
\label{sec:large-kernel-architecture}
Following the above designs and observations, we now describe the architectures of
our large-kernel 3D CNNs. We readily replace the plain 3D submanifold sparse convolution~\cite{submanifold-sparse-conv-v2} layers with our SW-LK Conv in existing 3D backbones for semantic segmentation and object detection.

\vspace{0.5em}
\noindent
\textbf{3D Semantic Segmentation}
The architectures of LargeKernel3D for 3D semantic segmentation resemble MinkowskiNet~\cite{minkowskinet}, including 1 stem layer and 8 stages. 
The stem layer is a 5$\times$5$\times$5 submanifold convolution. The first 4 stages account for down-sampling and the last 4 stages are for up-sampling. As for the U-Net structure, the first 4 features are concatenated into the up-sampling stages. We replace the original layers in MinkowskiNet-34~\cite{minkowskinet} by SW-LK Convs, where spatial-wise 7$\times$7$\times$7 convolutions is grouped into 3$\times$3$\times$3 splits. By default, other hyper-parameters, including channels and block numbers, follow MinkowskiNet-34~\cite{minkowskinet}.

\vspace{0.5em}
\noindent
\textbf{3D Object Detection} The typical backbone networks in 3D object detectors~\cite{pvrcnn, voxel-rcnn, centerpoint} consist of 1 stem layer and 4 stages. A 3$\times$3$\times$3 submanifold convolution~\cite{submanifold-sparse-conv-v2} layer serves as the stem. In each stage, except for the first one, there is a sparse convolutional layer with stride 2 for down-sampling. We substitute other plain blocks with SW-LK Convs. For models on nuScenes dataset, we find that 7$\times$7$\times$7 kernel sizes are enough. They both use the 3$\times$3$\times$3 spatial-wise groups.
For models on Waymo dataset, kernel sizes are scalable to 17$\times$17$\times$17. We remain the last stage of backbone network as the original, because receptive fields have already been sufficiently enlarged in the front stages. We keep these hyper-parameters unchanged by the default settings in baseline detectors~\cite{voxel-rcnn, centerpoint}. We include these detailed numbers in {\em appendix}. 

\section{Experiments}
\label{sec:experiments}
\subsection{Experimental Setting}
\vspace{0.5em}
\noindent
\textbf{3D Semantic Segmentation}
ScanNetv2~\cite{scannet} is a large-scale benchmark in 3D semantic segmentation. It contains 1,201 indoor RGB-D scenes for training, 312 scenes for validation, and 100 scenes for testing. Semantic labels in 20 categories are annotated. We train our models with the default settings in MinkowskiNet~\cite{minkowskinet}, including training hyper-parameters and data augmentations. We evaluate models in the mean Intersection over Union~(mIoU) metric. Detailed experimental settings, including network structures and training details, are listed in the {\em appendix}.

\begin{table}[t]
\begin{center}
\caption{ {Ablations on various techniques and kernel sizes on MinkowskiNet-34 and ScanNetv2. LN denotes the layer normalization. DW Conv denotes the depth-wise convolution.}}
\resizebox{\linewidth}{!}{
\begin{tabular}{|c|c|c|c|cc|c|}
\hline
Kernel  & Baseline  & +GELU & +LN & +DW Conv  & +PW  & Ours \\ \hline
 3$\times$3$\times$3 & 71.7            & 70.2   & 66.8         & 70.6   & 70.7           & -             \\ 
 7$\times$7$\times$7 & 68.6            & 68.4   & 65.0         & 68.7    & 68.7          & \textbf{73.5}          \\ \hline
\end{tabular}}
\label{tab:ablation-scannet}
\end{center}
\end{table}
\begin{table}[t]
\begin{center}
\caption{Ablations on group convolution on ScanNetv2.}
\resizebox{\linewidth}{!}{
\begin{tabular}{|c|c|ccccc|c|}
\hline
Kernel  & Baseline  & Group 2 & 4 & 8 & 16  &  DW & Ours \\ \hline
3x3x3 & 71.7 & 71.5 & 71.2 & 70.7 & 70.6 & 70.6 & - \\ \hline
7x7x7 & 68.6 & 68.8 & 69.2 & 68.8 & 68.7 & 68.7 & 73.5 \\ \hline
\end{tabular}}
\label{tab:group-dw-conv}
\end{center}
\end{table}
\begin{table}[t]
\begin{center}
\caption{ {Comparisons with relevant convolutional schemes on ScanNetv2. Baseline is a MinkowskiNet-34 network. All models listed are trained in the same training hyper-parameters.}}
\resizebox{\linewidth}{!}{
\begin{tabular}{|l|rrr|c|}
\hline
Method              & Params & FLOPs  & Latency & mIoU \\ \hline
Baseline & 37.9 M   & 182.8 G & 108 ms    & 71.7 \\
+ Kernel 5$\times$5$\times$5 & 170.3M  & 537.5 G & 212 ms    & 70.7 \\
+ Kernel 7$\times$7$\times$7 & 465.0M  & 1089.5 G & 487 ms    & 68.6 \\ \hline
Dilate conv       & 37.9M  & 100.1 G & 98 ms  & 64.6 \\
Pooling + Dilate      & 37.9M  & 183.2 G & 115 ms    & nan  \\
Spatial group~\cite{sgc}  & 37.9M  & 127.2 G & 96 ms & 70.0 \\
Deformable conv    & 42.5M  & 250.1 G & 238 ms    & 72.0 \\ \hline
LargeKernel3D-T & 38.4M  & 171.4 G &  111 ms & 72.8 \\
LargeKernel3D                & 40.2M  & 240.0 G & 145 ms & \textbf{73.5} \\ \hline
\end{tabular}}
\label{tab:comparisons-convolutions-scannet}
\end{center}
\end{table}
\begin{table}[t]
\begin{center}
\caption{Results on other backbone networks andScanNetv2.}
\begin{tabular}{|c|ccc|}
\hline
Model & Baseline & +Ours & $\Delta$ \\ \hline
MinkowskiNet-14~\cite{minkowskinet} & 67.0 & 69.7 & +2.7 \\ 
MinkowskiNet-18~\cite{minkowskinet} & 69.5 & 71.5 & +2.0 \\ \hline
SparseConvNet~\cite{sparseconvnet} & 69.3 & 71.2 & +1.9 \\ \hline
\end{tabular}
\label{tab:other-sparse-cnns}
\end{center}
\end{table}

\vspace{0.5em}
\noindent
\textbf{3D Object Detection}
nuScenes, and Waymo are all popular 3D object detection benchmarks.
nuScenes is a large dataset and contains 1,000 driving sequences in total. Among them, there are 700 scenes split for training, 150 scenes for validation, and 150 scenes for testing. It contains LIDAR, camera, and radar sources with a complete 360$^{\mathrm{o}}$ environment. We evaluate our methods on both LIDAR-only and LIDAR-RGB fusion settings. The main evaluation metrics are mAP and nuScenes detection score~(NDS).  {Waymo~\cite{waymo} dataset also contains 1,000 sequences in total, with 798 for training and 202 for validation. For Waymo dataset, we use 1/5 training data for the ablation study and full training set for main results.
Results in Waymo are evaluated on difficulty LEVEL\_1 and LEVEL\_2 objects.} In experiments, we validate our networks on state-of-the-art detection framework of CenterPoint~\cite{centerpoint} on nuScenes~\cite{nuscenes} and Waymo~\cite{waymo}.

\begin{table}[t]
\begin{center}
\caption{{Effects of the spatial-wise partition manners on SW-LK Conv and ScanNetv2. Center group means the size of central area. Shifting is to translate the center position by 1 in all 3 dimensions.}}
\begin{tabular}{|c|cccc|}
\hline
Center group  &  1$\times$1$\times$1 &  1$\times$1$\times$1 & 3$\times$3$\times$3 &  5$\times$5$\times$5  \\ \hline
Shifting & \xmark  & \cmark & \xmark  & \xmark \\ \hline \hline
mIoU & \textbf{73.5}   &  66.8  & 70.6    &   68.2   \\ \hline
\end{tabular}
\label{tab:group-partition-manner}
\end{center}
\end{table}
\begin{table}[t]
\begin{center}
\caption{{Improvement over various kernel sizes on LargeKernel3D upon CenterPoint, training on  $1/5$ Waymo training set and evaluate on the validation set. $\Delta$ means the improvement from the kernel 3 baseline. L1 and L2 means LEVEL\_1 and LEVEL\_2 mAP.}}
\resizebox{\linewidth}{!}{
\begin{tabular}{|c|cc|cc|cc|}
\hline
\multirow{2}{*}{\small Kernel} & \multicolumn{2}{c|}{Vehicle} & \multicolumn{2}{c|}{Pedestrian} & \multicolumn{2}{c|}{Cyclist} \\
                        & L1            & L2            & L1              & L2              & L1             & L2            \\ \hline
Baseline  & 70.9                                                   & 62.9                                                   & 71.5                                                   & 63.5                                                   & 69.1                                                   & 66.5                                                   \\
7$\times$7$\times$7  & 71.9                                                   & 63.8                                                   & 71.7                                                   & 63.7                                                   & 70.4                                                   & 67.8                                                   \\
11$\times$11$\times$11 & 72.2                                                   & 64.2                                                   & 71.8                                                   & 63.9                                                   & 70.2                                                   & 68.3                                                   \\
15$\times$15$\times$15 & 72.7                                                   & 64.6                                                   & 73.8                                                   & 65.8                                                   & 70.8                                                   & 68.2                                                   \\
17$\times$17$\times$17 & \textbf{73.2} & \textbf{65.1} & \textbf{74.1} & \textbf{66.0} & \textbf{71.0} & \textbf{68.1} \\ 
$\Delta$ & 2.3$\uparrow$ & 2.2$\uparrow$ &  2.6$\uparrow$ &  2.5$\uparrow$ &  1.9$\uparrow$ & 1.6$\uparrow$ \\ \hline
\end{tabular}}
\label{tab:waymo}
\end{center}
\end{table}
\begin{table}[t]
\begin{center}
\caption{ {Effects of position embedding~(PE) in the mAP of vehicle detection on Waymo validation set.}}
\begin{tabular}{|c|c|c|cc|}
\hline
Kernel                  & PE & Latency & LEVEL\_1                         & LEVEL\_2                         \\ \hline
3$\times$3$\times$3                   &      -              & 109 ms   & 70.9                           & 62.9                           \\ \hline
\multirow{2}{*}{7$\times$7$\times$7}  &       \xmark             &    120 ms     &      71.6                          &         63.4                       \\
                        &       \cmark             & 122 ms   & 71.9                           & 63.8                           \\ \hline
\multirow{2}{*}{17$\times$17$\times$17} &       \xmark             &    158 ms     &       71.5                         &         63.3                       \\
                        &       \cmark             & 164 ms   & 73.2 & 65.1 \\ \hline
\end{tabular}
\label{tab:effects-position-embedding}
\end{center}
\end{table}

\begin{table*}[t]
\begin{center}
\caption{Comparison with other methods on nuScenes {\em test} split. $^{\ddagger}$ means flipping and rotation testing-time augmentations.}
\resizebox{\linewidth}{!}{
\begin{tabular}{|l|cc|cccccccccc|}
\hline
Method          & NDS  & mAP  & Car                  & Truck                & Bus                  & Trailer              & C.V.                   & Ped                  & Mot                  & Byc                  & T.C.                   & Bar                   \\ \hline \hline
PointPillars~\cite{pointpillars}                     & 45.3 & 30.5 & 68.4 & 23.0 & 28.2 & 23.4 & 4.1 & 59.7 & 27.4 & 1.1 & 30.8 & 38.9 \\ 
3DSSD~\cite{3dssd}                         & 56.4 & 42.6 &  81.2 & 47.2 & 61.4 & 30.5 & 12.6 & 70.2 & 36.0 & 8.6 & 31.1 & 47.9 \\ 
CBGS~\cite{cbgs}                        & 63.3 & 52.8 & 81.1 & 48.5 & 54.9 & 42.9 & 10.5 & 80.1 & 51.5 & 22.3 & 70.9 & 65.7 \\ 
CenterPoint~\cite{centerpoint}                         & 65.5 & 58.0 & 84.6 & 51.0 & 60.2 & 53.2 & 17.5 & 83.4 & 53.7 & 28.7 & 76.7 & 70.9 \\ 
HotSpotNet~\cite{hotspotnet}                   & 66.0 & 59.3 & 83.1 & 50.9 & 56.4 & 53.3 & 23.0 & 81.3 & 63.5 & 36.6 & 73.0 & 71.6 \\
CVCNET~\cite{cvcnet}                  & 66.6 & 58.2 & 82.6 & 49.5 & 59.4 & 51.1 & 16.2 & 83.0 & 61.8 & 38.8 & 69.7 & 69.7 \\ 
UVTR-L~\cite{uvtr} & 69.7 & 63.9  & 86.3 & 52.2 & 62.8 & 59.7 & 33.7 & 84.5 & 68.8 & 41.1 & 74.7 & 74.9 \\ 
VISTA~\cite{vista} & 69.8 & 63.0 & 84.4 & 55.1 & 63.7 & 54.2 & 25.1 & 82.8 & 70.0 & 45.4 & 78.5 & 71.4 \\
Focals Conv~\cite{focal-sparse-conv}  & 70.0 & 63.8 & 86.7 & 56.3 & 67.7 & 59.5 & 23.8 & 87.5 & 64.5 & 36.3 & 81.4 & 74.1 \\ 
TransFusion-L~\cite{transfusion}                      & 70.2 & 65.5 & 86.2 & 56.7 & 66.3 & 58.8 & 28.2 & 86.1 & 68.3 & 44.2 & 82.0 & 78.2 \\ 
\hline \hline
 {LargeKernel3D}                      & 70.6 & 65.4 & 85.5 & 53.8 & 64.4 & 59.5 & 29.7 & 85.9 & 72.7 & 46.8 & 79.9 & 75.5 \\ 
 {LargeKernel3D $^{\ddagger}$}                      & \textbf{72.8} & \textbf{68.7} & 86.7 & 58.5 & 67.7 & 62.7 & 31.9 & 88.5 & 77.1 & 54.9 & 82.3 & 76.6 \\ 
 {LargeKernel3D-F$^{\ddagger}$}                      & \textbf{74.2} & \textbf{71.2} & 87.7 & 60.1 & 69.3 & 66.0 & 34.3 & 89.4 & 81.3 & 60.2 & 86.7 & 77.4 \\ 
\hline
\end{tabular}}
\label{tab:nuscenes-test}
\end{center}
\end{table*}

\subsection{Ablation Studies}
\vspace{0.5em}
\noindent
\textbf{Usefulness of 2D CNN Techniques}
We first validate the techniques that have proven effective on 2D CNNs, including layer normalization~\cite{layer-norm}, depth-wise convolution~\cite{mobilenet, convnext}, and GeLU~\cite{gelus}. We conduct experiments on MinkowskiNet-34~\cite{minkowskinet} on the ScanNet~\cite{scannet} semantic segmentation dataset, as shown in Tab.~\ref{tab:ablation-scannet}. Kernel 7$\times$7$\times$7 means modifying kernel sizes in all stages to 7. All these techniques bring no obvious benefit to baseline networks. In contrast, SW-LK Conv introduces clear improvement upon these strong networks. We also show the effects of group convolution with different group numbers in Tab.~\ref{tab:group-dw-conv}.

\vspace{0.5em}
\noindent
\textbf{Comparisons with Relevant Schemes}
We compare both performance and efficiency in Tab.~\ref{tab:comparisons-convolutions-scannet} using various existing schemes to enlarge receptive fields. Dilated Conv~\cite{deeplabv3} contains 3$\times$3$\times$3 kernel parameters. It dilates to 7$\times$7$\times$7 spatial sizes. Pooling + Dilated Conv inserts a 3D average pooling layer before the dilated convolutions to enlarge receptive fields. Spatial group conv is implemented, following the settings in the original paper~\cite{sgc}. We extend Deformable Conv~\cite{deformableconv} to 3D offset prediction, following~\cite{kpconv}.
Latencies are measured on an NVIDIA 2080ti GPU. It shows that none of these related schemes present better performance than the baseline. 
One reason for the low performance of dilation and deformable convolutions is that they tend to factorize the original dense kernels into sparse ones, which cause more information loss when applied to sparse 3D features than their 2D counterparts. Our LargeKernel3D achieves the best performance among these methods. In addition, we also build a tiny version, LargeKernel3D-T, which has half channel numbers in the last two stages. It introduces almost no additional computation overhead to baseline, but still has notable performance improvement. Except for the MinkowskiNet-34, we also show our effects on other backbone networks in Tab.~\ref{tab:other-sparse-cnns}.

\begin{table}[t]
\centering
        \caption{Comparison on ScanNetv2 mIoU on 3D semantic segmentation. $^{\dagger}$ Sliding-window testing.}
        \begin{tabular}{|l|ccc|}
        \hline
        Method         & Latency       & {\em val} & \textbf{\em test} \\ \hline
        PointCNN~\cite{pointcnn} & - & - & 45.8 \\
        PointNet++~\cite{pointnet++} & - & 53.5 & 55.7 \\
        RandLA-Net~\cite{randla-net} & - & - & 64.5 \\
        PointConv~\cite{pointconv} & 80 ms & 61.0 & 66.6 \\
        PointASNL~\cite{pointasnl} & - & 63.5 & 66.6 \\
        KPConv~\cite{kpconv} & - & 69.2 & 68.6 \\
        FusionNet~\cite{fusionnet} & - & - & 68.8 \\ 
        Point Transformer$^{\dagger}$~\cite{point-transformer} & 1012 ms & 70.6 & - \\ 
        Fast Point Transformer~\cite{fast-point-transformer} & 298 ms & 72.1 & - \\
        SparseConvNet~\cite{sparseconvnet} & - & 69.3 & 72.5 \\ 
        Stratified Transformer$^{\dagger}$~\cite{stratified-transformer} & 1624 ms & 74.3 & 73.7 \\ 
        \hline \hline
        MinkowskiNet-34         & 108 ms & 71.7 &  \\
        + LargeKernel3D & 145 & 73.5 & \textbf{73.9} \\ \hline
        \end{tabular}
        \label{tab:comparison-scannetv2}
    \hfill
\end{table}

\vspace{0.5em}
\noindent
\textbf{Spatial-wise Group Partition Manner}
We ablate the effects of different spatial-wise group partition manners in Tab.~\ref{tab:group-partition-manner}. We study the size of central groups and center shifting. This ablation study is conducted on the LargeKernel3D and the Scannetv2 dataset. We find that both large central sizes and center shifting hurt the performance. As the center sizes enlarge from 1$\times$1$\times$1 to 5$\times$5$\times$5, the accuracy sharply drops. Because submanifold sparse convolution~\cite{submanifold-sparse-conv-v2} always aligns each individual voxel feature at its kernel center, we suppose that central features should maintain distinct and clean. Therefore, we keep the center size as 1$\times$1$\times$1 without shifting, as a default setting.

\vspace{0.5em}
\noindent
\textbf{Further Enlarging Kernel Sizes}
We show that LargeKernel3D is scalable to kernel sizes 17$\times$17$\times$17 on the outdoor dataset Waymo~\cite{waymo} in Tab.~\ref{tab:waymo} with baseline kernel 3$\times$3$\times$3. Models in this ablation are trained on $1/5$ training data and evaluated on the full validation set. Results with full training data are compared in Tab.~\ref{tab:waymo-val}. It shows that the mAP on all categories increases by around 2\%. This setting demonstrates that a large-scale dataset is a key to unlocking the potential and capability of large kernel methods. In addition, we also try to extend the kernel size to $19\times19\times19$, but obtain no further gains.

\begin{table}[t]
\begin{center}
\caption{Performance of vehicle detection on the Waymo {\em val} split. All models listed take LIDAR-only input and single frames, without any test-time augmentations or model ensemble.}
\resizebox{\linewidth}{!}{
\begin{tabular}{|l|cc|}
\hline
Method                        &   \makecell[c]{LEVEL$\_$1\\3D AP / APH}    & \makecell[c]{LEVEL$\_2$\\3D AP / APH}   \\ \hline \hline
VoTr-SSD~\cite{voxeltransformer}   & 68.99 / 68.39 & 60.22 / 59.69  \\
PointPillars~\cite{centerpoint}   & 72.08 / 71.53 & 63.55 / 63.06  \\
SECOND~\cite{second}   & 72.27 / 71.69 & 63.85 / 63.33  \\
RangeDet~\cite{rangedet}   & 72.85 / 72.33 & 64.03 / 63.57  \\
SST~\cite{single-stride-transformer}   & 74.22 / 73.77 & 65.47 / 65.07  \\
VoTr-TSD~\cite{voxeltransformer}   & 74.95 / 74.25 & 65.91 / 65.29  \\
RSN~\cite{rsn}   & 75.1 / 74.6 & 66.0 / 65.5  \\
Voxel RCNN~\cite{voxel-rcnn}   & 75.59 / ------- & 66.59 / -------  \\
LIDAR-RCNN~\cite{lidar-rcnn}   & 76.0 / 75.5 & 68.3 / 67.9  \\
SST~\cite{single-stride-transformer}   & 76.22 / 75.79 & 68.04 / 67.64  \\
Pyramid RCNN~\cite{pyramid-rcnn}   & 76.30 / 75.68 & 67.23 / 66.68  \\
Part-A2-Net~\cite{part-a2}   & 77.05 / 76.51 & 68.47 / 67.14  \\
PV-RCNN~\cite{pvrcnn}   & 77.51 / 76.89 & 68.98 / 68.41  \\
SWFormer~\cite{swformer}                                       & 69.2 / 68.8    & 80.9 / 72.7 \\ \hline
CenterPoint~\cite{centerpoint}   & 76.59 / 76.05 & 68.85 / 68.35  \\
+ LargeKernel3D   &   \textbf{78.07} / \textbf{77.61}  & \textbf{69.81} / \textbf{69.38}  \\ \hline
\end{tabular}
}
\label{tab:waymo-val}
\end{center}
\end{table}

\vspace{0.5em}
\noindent
\textbf{Effects of Position Embedding}
In LargeKernel3D, position embedding is introduced to relieve the feature-blurring issue caused by weight-sharing. We ablate its effects in Tab.~\ref{tab:effects-position-embedding}, in terms of both latency and performance. Position embedding introduces limited latency overhead on both kernel 7$\times$7$\times$7 and 17$\times$17$\times$17 models. In terms of performance, its effects on the 7$\times$7$\times$7 model are a bit marginal, but essential on the 17$\times$17$\times$17 model, which has larger kernel sizes and suffers from feature blurring.

\vspace{0.5em}
\noindent
\textbf{Shrinking Kernels During Inference}
For the spatial-wise partition convolution, we can shrink kernels into small ones during inference. As shown in Tab.~\ref{tab:comparisons-convolutions-scannet}, the latency of LargeKernel3D is 145ms on ScanNetv2. It degrades to 514ms with the same 73.5\% mIoU, if we disable shrinking.

\subsection{Main Results}
\vspace{0.5em}
\noindent
\textbf{3D Semantic Segmentation}
We make comparisons with other 3D semantic segmentation methods on the ScanNetv2 dataset in Tab.~\ref{tab:comparison-scannetv2}. 
Our method surpasses others on {\em test} split. MinkowskiNet~\cite{minkowskinet} is already a state-of-the-art method in ScanNetv2. Our SW-LK Conv further improves its performance. Our method is also superior to transformer-based methods~\cite{point-transformer,fast-point-transformer}. Stratified Transformer~\cite{stratified-transformer} uses a sliding-window testing technique$^{\dagger}$ and splits integrated scenes into overlapped parts. The method tests each part one by one and then ensembles results together, taking hours on evaluation. Moreover, our results are better on the {\em test} split.

\vspace{0.5em}
\noindent
\textbf{3D Object Detection} We compare our LargeKernel3D upon CenterPoint~\cite{centerpoint} with previous methods on the {\em test} split of the nuScenes~\cite{nuscenes} dataset in Tab.~\ref{tab:nuscenes-test}. LargeKernel3D improves CenterPoint~\cite{centerpoint} to 70.6\% and 72.8\% NDS, with and without test augmentation, both outperforming other LIDAR methods. The multi-modal modal LargeKernel3D-F further improves to 74.2\% NDS and 71.2\% mAP. The fusion module directly follows a simple painting-based manner~\cite{focal-sparse-conv}.
Results on the {\em val} split is shown in the appendix.

We also show the effectiveness of LargeKernel3D model on the vehicle detection of Waymo~\cite{waymo} validation set in Tab.~\ref{tab:waymo-val}. The kernel size of LargeKernel3D is $17\times$17$\times$17. LargeKernel3D improves the vehicle performance of CenterPoint~\cite{centerpoint} by around 1.5\% LEVEL\_1 AP / APH. Models listed  take single-frame LIDAR data without any test-time augmentations or ensemble. It also performs better than other methods. We use CenterPoint~\cite{centerpoint} to keep consistency to the nuScenes benchmark. Note that LargeKernel3D has the potential to achieve further performance if equipped with stronger detection heads~\cite{pvrcnn}. 

\section{Conclusion and Discussion}
\label{sec:conclusion}
We have studied large-kernel designs for 3D convolutional networks, which have essential differences from the solutions in 2D CNNs. We present {spatial-wise partition convolution~(SW Conv)} that is specifically designed for 3D large kernels. It effectively resolves the efficiency and optimization issues in plain 3D large-kernel CNNs. Based on this design, we further propose the SW-LK Conv and the corresponding LargeKernel3D for 3D semantic segmentation and object detection. Our 3D large-kernel networks achieve decent improvement on both semantic segmentation and object detection tasks. 
For the first time, we show that 3D large kernels can be realized efficiently and effectively. We expect our findings to advance further development of 3D networks. Additional experiments and limitation analysis are provided in the {\em appendix}.

\vspace{0.35em}
\noindent
\textbf{Limitations}
LargeKernel3D mainly relies on hand-designed spatial kernel sizes in 3D semantic segmentation and object detection benchmarks. These sizes might be sub-optimal for other datasets or tasks, depending on the overall scene sizes and data sparsity. Other search-based techniques~\cite{spvnas} might be helpful, which we will try latter.

\vspace{0.35em}
\noindent
\textbf{Boarder impacts}
The proposed method can serve backbone network for various 3D tasks, which might include tasks or datasets involving negative societal impacts.

{\small
\bibliographystyle{ieee_fullname}
\bibliography{egbib}

\begin{thebibliography}{10}\itemsep=-1pt

\bibitem{layer-norm}
Lei~Jimmy Ba, Jamie~Ryan Kiros, and Geoffrey~E. Hinton.
\newblock Layer normalization.
\newblock {\em CoRR}, abs/1607.06450, 2016.

\bibitem{transfusion}
Xuyang Bai, Zeyu Hu, Xinge Zhu, Qingqiu Huang, Yilun Chen, Hongbo Fu, and
  Chiew{-}Lan Tai.
\newblock Transfusion: Robust lidar-camera fusion for 3d object detection with
  transformers.
\newblock {\em CoRR}, abs/2203.11496, 2022.

\bibitem{bello2019attention}
Irwan Bello, Barret Zoph, Ashish Vaswani, Jonathon Shlens, and Quoc~V Le.
\newblock Attention augmented convolutional networks.
\newblock In {\em Proceedings of the IEEE/CVF international conference on
  computer vision}, pages 3286--3295, 2019.

\bibitem{nuscenes}
Holger Caesar, Varun Bankiti, Alex~H. Lang, Sourabh Vora, Venice~Erin Liong,
  Qiang Xu, Anush Krishnan, Yu Pan, Giancarlo Baldan, and Oscar Beijbom.
\newblock nuscenes: {A} multimodal dataset for autonomous driving.
\newblock In {\em CVPR}, pages 11618--11628, 2020.

\bibitem{deeplabv3}
Liang{-}Chieh Chen, George Papandreou, Florian Schroff, and Hartwig Adam.
\newblock Rethinking atrous convolution for semantic image segmentation.
\newblock {\em CoRR}, abs/1706.05587, 2017.

\bibitem{cvcnet}
Qi Chen, Lin Sun, Ernest Cheung, and Alan~L. Yuille.
\newblock Every view counts: Cross-view consistency in 3d object detection with
  hybrid-cylindrical-spherical voxelization.
\newblock In {\em NeurIPS}, 2020.

\bibitem{hotspotnet}
Qi Chen, Lin Sun, Zhixin Wang, Kui Jia, and Alan~L. Yuille.
\newblock Object as hotspots: An anchor-free 3d object detection approach via
  firing of hotspots.
\newblock In {\em ECCV}, volume 12366, pages 68--84, 2020.

\bibitem{focal-sparse-conv}
Yukang Chen, Yanwei Li, Xiangyu Zhang, Jian Sun, and Jiaya Jia.
\newblock Focal sparse convolutional networks for 3d object detection.
\newblock {\em CoRR}, abs/2204.12463, 2022.

\bibitem{minkowskinet}
Christopher~B. Choy, JunYoung Gwak, and Silvio Savarese.
\newblock 4d spatio-temporal convnets: Minkowski convolutional neural networks.
\newblock In {\em CVPR}, pages 3075--3084, 2019.

\bibitem{chu2021icm}
Ruihang Chu, Yukang Chen, Tao Kong, Lu Qi, and Lei Li.
\newblock Icm-3d: Instantiated category modeling for 3d instance segmentation.
\newblock {\em IEEE Robotics and Automation Letters}, 7(1):57--64, 2021.

\bibitem{chu2022twist}
Ruihang Chu, Xiaoqing Ye, Zhengzhe Liu, Xiao Tan, Xiaojuan Qi, Chi-Wing Fu, and
  Jiaya Jia.
\newblock Twist: Two-way inter-label self-training for semi-supervised 3d
  instance segmentation.
\newblock In {\em Proceedings of the IEEE/CVF conference on computer vision and
  pattern recognition}, pages 1100--1109, 2022.

\bibitem{chu2021conditional}
Xiangxiang Chu, Zhi Tian, Bo Zhang, Xinlong Wang, Xiaolin Wei, Huaxia Xia, and
  Chunhua Shen.
\newblock Conditional positional encodings for vision transformers.
\newblock {\em arXiv preprint arXiv:2102.10882}, 2021.

\bibitem{scannet}
Angela Dai, Angel~X. Chang, Manolis Savva, Maciej Halber, Thomas Funkhouser,
  and Matthias Nie{\ss}ner.
\newblock Scannet: Richly-annotated 3d reconstructions of indoor scenes.
\newblock In {\em CVPR}, 2017.

\bibitem{deformableconv}
Jifeng Dai, Haozhi Qi, Yuwen Xiong, Yi Li, Guodong Zhang, Han Hu, and Yichen
  Wei.
\newblock Deformable convolutional networks.
\newblock In {\em ICCV}, pages 764--773, 2017.

\bibitem{imagenet}
Jia Deng, Wei Dong, Richard Socher, Li{-}Jia Li, Kai Li, and Li Fei{-}Fei.
\newblock Imagenet: {A} large-scale hierarchical image database.
\newblock In {\em CVPR}, pages 248--255, 2009.

\bibitem{voxel-rcnn}
Jiajun Deng, Shaoshuai Shi, Peiwei Li, Wengang Zhou, Yanyong Zhang, and
  Houqiang Li.
\newblock Voxel {R-CNN:} towards high performance voxel-based 3d object
  detection.
\newblock In {\em AAAI}, pages 1201--1209, 2021.

\bibitem{vista}
Shengheng Deng, Zhihao Liang, Lin Sun, and Kui Jia.
\newblock {VISTA:} boosting 3d object detection via dual cross-view spatial
  attention.
\newblock In {\em CVPR}, pages 8438--8447, 2022.

\bibitem{large-kernel}
Xiaohan Ding, Xiangyu Zhang, Yizhuang Zhou, Jungong Han, Guiguang Ding, and
  Jian Sun.
\newblock Scaling up your kernels to 31x31: Revisiting large kernel design in
  cnns.
\newblock In {\em CVPR}, pages 11963--11975, 2022.

\bibitem{cswin-transformer}
Xiaoyi Dong, Jianmin Bao, Dongdong Chen, Weiming Zhang, Nenghai Yu, Lu Yuan,
  Dong Chen, and Baining Guo.
\newblock Cswin transformer: {A} general vision transformer backbone with
  cross-shaped windows.
\newblock {\em CoRR}, abs/2107.00652, 2021.

\bibitem{vit}
Alexey Dosovitskiy, Lucas Beyer, Alexander Kolesnikov, Dirk Weissenborn,
  Xiaohua Zhai, Thomas Unterthiner, Mostafa Dehghani, Matthias Minderer, Georg
  Heigold, Sylvain Gelly, Jakob Uszkoreit, and Neil Houlsby.
\newblock An image is worth 16x16 words: Transformers for image recognition at
  scale.
\newblock In {\em ICLR}, 2021.

\bibitem{single-stride}
Lue Fan, Ziqi Pang, Tianyuan Zhang, Yu{-}Xiong Wang, Hang Zhao, Feng Wang,
  Naiyan Wang, and Zhaoxiang Zhang.
\newblock Embracing single stride 3d object detector with sparse transformer.
\newblock {\em CoRR}, abs/2112.06375, 2021.

\bibitem{single-stride-transformer}
Lue Fan, Ziqi Pang, Tianyuan Zhang, Yu{-}Xiong Wang, Hang Zhao, Feng Wang,
  Naiyan Wang, and Zhaoxiang Zhang.
\newblock Embracing single stride 3d object detector with sparse transformer.
\newblock In {\em CVPR}, pages 8448--8458, 2022.

\bibitem{rangedet}
Lue Fan, Xuan Xiong, Feng Wang, Naiyan Wang, and Zhaoxiang Zhang.
\newblock Rangedet: In defense of range view for lidar-based 3d object
  detection.
\newblock In {\em ICCV}, pages 2898--2907, 2021.

\bibitem{sparseconvnet}
Benjamin Graham, Martin Engelcke, and Laurens van~der Maaten.
\newblock 3d semantic segmentation with submanifold sparse convolutional
  networks.
\newblock In {\em CVPR}, pages 9224--9232, 2018.

\bibitem{submanifold-sparse-conv-v2}
Benjamin Graham, Martin Engelcke, and Laurens van~der Maaten.
\newblock 3d semantic segmentation with submanifold sparse convolutional
  networks.
\newblock In {\em CVPR}, pages 9224--9232, 2018.

\bibitem{m3detr}
Tianrui Guan, Jun Wang, Shiyi Lan, Rohan Chandra, Zuxuan Wu, Larry Davis, and
  Dinesh Manocha.
\newblock {M3DETR:} multi-representation, multi-scale, mutual-relation 3d
  object detection with transformers.
\newblock In {\em WACV}, pages 2293--2303, 2022.

\bibitem{connection-attention-dynamic-depthwise}
Qi Han, Zejia Fan, Qi Dai, Lei Sun, Ming-Ming Cheng, Jiaying Liu, and Jingdong
  Wang.
\newblock On the connection between local attention and dynamic depth-wise
  convolution.
\newblock In {\em ICLR}, 2021.

\bibitem{gelus}
Dan Hendrycks and Kevin Gimpel.
\newblock Gaussian error linear units (gelus).
\newblock {\em CoRR}, abs/1606.08415, 2016.

\bibitem{mobilenet}
Andrew~G. Howard, Menglong Zhu, Bo Chen, Dmitry Kalenichenko, Weijun Wang,
  Tobias Weyand, Marco Andreetto, and Hartwig Adam.
\newblock Mobilenets: Efficient convolutional neural networks for mobile vision
  applications.
\newblock {\em CoRR}, abs/1704.04861, 2017.

\bibitem{randla-net}
Qingyong Hu, Bo Yang, Linhai Xie, Stefano Rosa, Yulan Guo, Zhihua Wang, Niki
  Trigoni, and Andrew Markham.
\newblock Randla-net: Efficient semantic segmentation of large-scale point
  clouds.
\newblock In {\em CVPR}, pages 11105--11114, 2020.

\bibitem{active-conv}
Yunho Jeon and Junmo Kim.
\newblock Active convolution: Learning the shape of convolution for image
  classification.
\newblock In {\em CVPR}, pages 1846--1854, 2017.

\bibitem{jiang2021guided}
Li Jiang, Shaoshuai Shi, Zhuotao Tian, Xin Lai, Shu Liu, Chi-Wing Fu, and Jiaya
  Jia.
\newblock Guided point contrastive learning for semi-supervised point cloud
  semantic segmentation.
\newblock In {\em CVPR}, pages 6423--6432, 2021.

\bibitem{lai2023spherical}
Xin Lai, Yukang Chen, Fanbin Lu, Jianhui Liu, and Jiaya Jia.
\newblock Spherical transformer for lidar-based 3d recognition.
\newblock In {\em CVPR}, 2023.

\bibitem{stratified-transformer}
Xin Lai, Jianhui Liu, Li Jiang, Liwei Wang, Hengshuang Zhao, Shu Liu, Xiaojuan
  Qi, and Jiaya Jia.
\newblock Stratified transformer for 3d point cloud segmentation.
\newblock {\em CoRR}, abs/2203.14508, 2022.

\bibitem{pointpillars}
Alex~H. Lang, Sourabh Vora, Holger Caesar, Lubing Zhou, Jiong Yang, and Oscar
  Beijbom.
\newblock Pointpillars: Fast encoders for object detection from point clouds.
\newblock In {\em CVPR}, pages 12697--12705, 2019.

\bibitem{pointcnn}
Yangyan Li, Rui Bu, Mingchao Sun, Wei Wu, Xinhan Di, and Baoquan Chen.
\newblock Pointcnn: Convolution on x-transformed points.
\newblock In {\em NeurIPS}, pages 828--838, 2018.

\bibitem{uvtr}
Yanwei Li, Yilun Chen, Xiaojuan Qi, Zeming Li, Jian Sun, and Jiaya Jia.
\newblock Unifying voxel-based representation with transformer for 3d object
  detection.
\newblock {\em CoRR}, abs/2206.00630, 2022.

\bibitem{li2021simultaneous}
Yiming Li, Tao Kong, Ruihang Chu, Yifeng Li, Peng Wang, and Lei Li.
\newblock Simultaneous semantic and collision learning for 6-dof grasp pose
  estimation.
\newblock In {\em 2021 IEEE/RSJ International Conference on Intelligent Robots
  and Systems (IROS)}, pages 3571--3578. IEEE, 2021.

\bibitem{lidar-rcnn}
Zhichao Li, Feng Wang, and Naiyan Wang.
\newblock Lidar {R-CNN:} an efficient and universal 3d object detector.
\newblock In {\em CVPR}, pages 7546--7555, 2021.

\bibitem{coco}
Tsung{-}Yi Lin, Michael Maire, Serge~J. Belongie, James Hays, Pietro Perona,
  Deva Ramanan, Piotr Doll{\'{a}}r, and C.~Lawrence Zitnick.
\newblock Microsoft {COCO:} common objects in context.
\newblock In {\em ECCV}, volume 8693, pages 740--755, 2014.

\bibitem{swin-transformer-v2}
Ze Liu, Han Hu, Yutong Lin, Zhuliang Yao, Zhenda Xie, Yixuan Wei, Jia Ning, Yue
  Cao, Zheng Zhang, Li Dong, Furu Wei, and Baining Guo.
\newblock Swin transformer {V2:} scaling up capacity and resolution.
\newblock {\em CoRR}, abs/2111.09883, 2021.

\bibitem{swin-transformer}
Ze Liu, Yutong Lin, Yue Cao, Han Hu, Yixuan Wei, Zheng Zhang, Stephen Lin, and
  Baining Guo.
\newblock Swin transformer: Hierarchical vision transformer using shifted
  windows.
\newblock In {\em ICCV}, pages 9992--10002, 2021.

\bibitem{convnext}
Zhuang Liu, Hanzi Mao, Chao{-}Yuan Wu, Christoph Feichtenhofer, Trevor Darrell,
  and Saining Xie.
\newblock A convnet for the 2020s.
\newblock {\em CoRR}, abs/2201.03545, 2022.

\bibitem{pyramid-rcnn}
Jiageng Mao, Minzhe Niu, Haoyue Bai, Xiaodan Liang, Hang Xu, and Chunjing Xu.
\newblock Pyramid {R-CNN:} towards better performance and adaptability for 3d
  object detection.
\newblock In {\em ICCV}, 2021.

\bibitem{voxeltransformer}
Jiageng Mao, Yujing Xue, Minzhe Niu, Haoyue Bai, Jiashi Feng, Xiaodan Liang,
  Hang Xu, and Chunjing Xu.
\newblock Voxel transformer for 3d object detection.
\newblock In {\em ICCV}, 2021.

\bibitem{fast-point-transformer}
Chunghyun Park, Yoonwoo Jeong, Minsu Cho, and Jaesik Park.
\newblock Fast point transformer.
\newblock {\em CoRR}, abs/2112.04702, 2021.

\bibitem{large-kernel-seg}
Chao Peng, Xiangyu Zhang, Gang Yu, Guiming Luo, and Jian Sun.
\newblock Large kernel matters - improve semantic segmentation by global
  convolutional network.
\newblock In {\em CVPR}, pages 1743--1751, 2017.

\bibitem{pointnet}
Charles~Ruizhongtai Qi, Hao Su, Kaichun Mo, and Leonidas~J. Guibas.
\newblock Pointnet: Deep learning on point sets for 3d classification and
  segmentation.
\newblock In {\em CVPR}, pages 77--85, 2017.

\bibitem{pointnet++}
Charles~Ruizhongtai Qi, Li Yi, Hao Su, and Leonidas~J. Guibas.
\newblock Pointnet++: Deep hierarchical feature learning on point sets in a
  metric space.
\newblock In {\em NeurIPS}, pages 5099--5108, 2017.

\bibitem{raffel2020exploring}
Colin Raffel, Noam Shazeer, Adam Roberts, Katherine Lee, Sharan Narang, Michael
  Matena, Yanqi Zhou, Wei Li, Peter~J Liu, et~al.
\newblock Exploring the limits of transfer learning with a unified text-to-text
  transformer.
\newblock {\em J. Mach. Learn. Res.}, 21(140):1--67, 2020.

\bibitem{vision-transformer-like-cnn}
Maithra Raghu, Thomas Unterthiner, Simon Kornblith, Chiyuan Zhang, and Alexey
  Dosovitskiy.
\newblock Do vision transformers see like convolutional neural networks?
\newblock In {\em NeurIPS}, pages 12116--12128, 2021.

\bibitem{global-filternet}
Yongming Rao, Wenliang Zhao, Zheng Zhu, Jiwen Lu, and Jie Zhou.
\newblock Global filter networks for image classification.
\newblock In {\em NeurIPS}, pages 980--993, 2021.

\bibitem{ckconv}
David~W. Romero, Anna Kuzina, Erik~J. Bekkers, Jakub~M. Tomczak, and Mark
  Hoogendoorn.
\newblock Ckconv: Continuous kernel convolution for sequential data.
\newblock {\em CoRR}, abs/2102.02611, 2021.

\bibitem{shaw2018self}
Peter Shaw, Jakob Uszkoreit, and Ashish Vaswani.
\newblock Self-attention with relative position representations.
\newblock {\em arXiv preprint arXiv:1803.02155}, 2018.

\bibitem{pvrcnn}
Shaoshuai Shi, Chaoxu Guo, Li Jiang, Zhe Wang, Jianping Shi, Xiaogang Wang, and
  Hongsheng Li.
\newblock {PV-RCNN:} point-voxel feature set abstraction for 3d object
  detection.
\newblock In {\em CVPR}, pages 10526--10535, 2020.

\bibitem{part-a2}
Shaoshuai Shi, Zhe Wang, Jianping Shi, Xiaogang Wang, and Hongsheng Li.
\newblock From points to parts: 3d object detection from point cloud with
  part-aware and part-aggregation network.
\newblock {\em T-PAMI}, 43(8):2647--2664, 2021.

\bibitem{waymo}
Pei Sun, Henrik Kretzschmar, Xerxes Dotiwalla, Aurelien Chouard, Vijaysai
  Patnaik, Paul Tsui, James Guo, Yin Zhou, Yuning Chai, Benjamin Caine, Vijay
  Vasudevan, Wei Han, Jiquan Ngiam, Hang Zhao, Aleksei Timofeev, Scott
  Ettinger, Maxim Krivokon, Amy Gao, Aditya Joshi, Yu Zhang, Jonathon Shlens,
  Zhifeng Chen, and Dragomir Anguelov.
\newblock Scalability in perception for autonomous driving: Waymo open dataset.
\newblock In {\em CVPR}, pages 2443--2451, 2020.

\bibitem{swformer}
Pei Sun, Mingxing Tan, Weiyue Wang, Chenxi Liu, Fei Xia, Zhaoqi Leng, and
  Dragomir Anguelov.
\newblock Swformer: Sparse window transformer for 3d object detection in point
  clouds.
\newblock {\em CoRR}, abs/2210.07372, 2022.

\bibitem{rsn}
Pei Sun, Weiyue Wang, Yuning Chai, Gamaleldin Elsayed, Alex Bewley, Xiao Zhang,
  Cristian Sminchisescu, and Dragomir Anguelov.
\newblock {RSN:} range sparse net for efficient, accurate lidar 3d object
  detection.
\newblock In {\em CVPR}, pages 5725--5734, 2021.

\bibitem{spvnas}
Haotian Tang, Zhijian Liu, Shengyu Zhao, Yujun Lin, Ji Lin, Hanrui Wang, and
  Song Han.
\newblock Searching efficient 3d architectures with sparse point-voxel
  convolution.
\newblock In {\em ECCV}, volume 12373, pages 685--702, 2020.

\bibitem{kpconv}
Hugues Thomas, Charles~R. Qi, Jean{-}Emmanuel Deschaud, Beatriz Marcotegui,
  Fran{\c{c}}ois Goulette, and Leonidas~J. Guibas.
\newblock Kpconv: Flexible and deformable convolution for point clouds.
\newblock In {\em ICCV}, pages 6410--6419, 2019.

\bibitem{patches-all-needed}
Asher Trockman and J.~Zico Kolter.
\newblock Patches are all you need?
\newblock {\em CoRR}, abs/2201.09792, 2022.

\bibitem{vaswani2017attention}
Ashish Vaswani, Noam Shazeer, Niki Parmar, Jakob Uszkoreit, Llion Jones,
  Aidan~N Gomez, {\L}ukasz Kaiser, and Illia Polosukhin.
\newblock Attention is all you need.
\newblock {\em Advances in neural information processing systems}, 30, 2017.

\bibitem{infofocus}
Jun Wang, Shiyi Lan, Mingfei Gao, and Larry~S. Davis.
\newblock Infofocus: 3d object detection for autonomous driving with dynamic
  information modeling.
\newblock In {\em ECCV}, volume 12355, pages 405--420, 2020.

\bibitem{wu2021rethinking}
Kan Wu, Houwen Peng, Minghao Chen, Jianlong Fu, and Hongyang Chao.
\newblock Rethinking and improving relative position encoding for vision
  transformer.
\newblock In {\em Proceedings of the IEEE/CVF International Conference on
  Computer Vision}, pages 10033--10041, 2021.

\bibitem{pointconv}
Wenxuan Wu, Zhongang Qi, and Fuxin Li.
\newblock Pointconv: Deep convolutional networks on 3d point clouds.
\newblock In {\em CVPR}, pages 9621--9630, 2019.

\bibitem{image-deconvolution}
Li Xu, Jimmy S.~J. Ren, Ce Liu, and Jiaya Jia.
\newblock Deep convolutional neural network for image deconvolution.
\newblock In {\em NeurIPS}, pages 1790--1798, 2014.

\bibitem{pointasnl}
Xu Yan, Chaoda Zheng, Zhen Li, Sheng Wang, and Shuguang Cui.
\newblock Pointasnl: Robust point clouds processing using nonlocal neural
  networks with adaptive sampling.
\newblock In {\em CVPR}, pages 5588--5597, 2020.

\bibitem{second}
Yan Yan, Yuxing Mao, and Bo Li.
\newblock {SECOND:} sparsely embedded convolutional detection.
\newblock {\em Sensors}, 18(10):3337, 2018.

\bibitem{focal-transformer}
Jianwei Yang, Chunyuan Li, Pengchuan Zhang, Xiyang Dai, Bin Xiao, Lu Yuan, and
  Jianfeng Gao.
\newblock Focal self-attention for local-global interactions in vision
  transformers.
\newblock {\em CoRR}, abs/2107.00641, 2021.

\bibitem{fast-part-c}
Tao Yang, Haokui Zhang, Wenze Hu, Changwen Chen, and Xiaoyu Wang.
\newblock Fast-parc: Position aware global kernel for convnets and vits.
\newblock {\em CoRR}, abs/2210.04020, 2022.

\bibitem{yang2019xlnet}
Zhilin Yang, Zihang Dai, Yiming Yang, Jaime Carbonell, Russ~R Salakhutdinov,
  and Quoc~V Le.
\newblock Xlnet: Generalized autoregressive pretraining for language
  understanding.
\newblock {\em Advances in neural information processing systems}, 32, 2019.

\bibitem{3dssd}
Zetong Yang, Yanan Sun, Shu Liu, and Jiaya Jia.
\newblock 3dssd: Point-based 3d single stage object detector.
\newblock In {\em CVPR}, pages 11037--11045, 2020.

\bibitem{centerpoint}
Tianwei Yin, Xingyi Zhou, and Philipp Kr{\"{a}}henb{\"{u}}hl.
\newblock Center-based 3d object detection and tracking.
\newblock In {\em CVPR}, pages 11784--11793, 2021.

\bibitem{atrous-conv}
Fisher Yu and Vladlen Koltun.
\newblock Multi-scale context aggregation by dilated convolutions.
\newblock In Yoshua Bengio and Yann LeCun, editors, {\em ICLR}, 2016.

\bibitem{fusionnet}
Feihu Zhang, Jin Fang, Benjamin~W. Wah, and Philip H.~S. Torr.
\newblock Deep fusionnet for point cloud semantic segmentation.
\newblock In {\em ECCV}, volume 12369, pages 644--663, 2020.

\bibitem{sgc}
Jiahui Zhang, Hao Zhao, Anbang Yao, Yurong Chen, Li Zhang, and Hongen Liao.
\newblock Efficient semantic scene completion network with spatial group
  convolution.
\newblock In {\em ECCV}, pages 733--749, 2018.

\bibitem{point-transformer}
Hengshuang Zhao, Li Jiang, Jiaya Jia, Philip H.~S. Torr, and Vladlen Koltun.
\newblock Point transformer.
\newblock In {\em ICCV}, pages 16239--16248, 2021.

\bibitem{cbgs}
Benjin Zhu, Zhengkai Jiang, Xiangxin Zhou, Zeming Li, and Gang Yu.
\newblock Class-balanced grouping and sampling for point cloud 3d object
  detection.
\newblock {\em CoRR}, abs/1908.09492, 2019.

\bibitem{deformableconvv2}
Xizhou Zhu, Han Hu, Stephen Lin, and Jifeng Dai.
\newblock Deformable convnets {V2:} more deformable, better results.
\newblock In {\em CVPR}, pages 9308--9316, 2019.

\end{thebibliography}
}

\newpage
\appendix
\captionsetup[table]{labelformat={default},labelsep=period,name={Table S -}}
\captionsetup[figure]{labelformat={default},labelsep=period,name={Figure S -}}

\section*{Appendix}
\begin{figure}[t]
\begin{center}
   \includegraphics[width=0.8\linewidth]{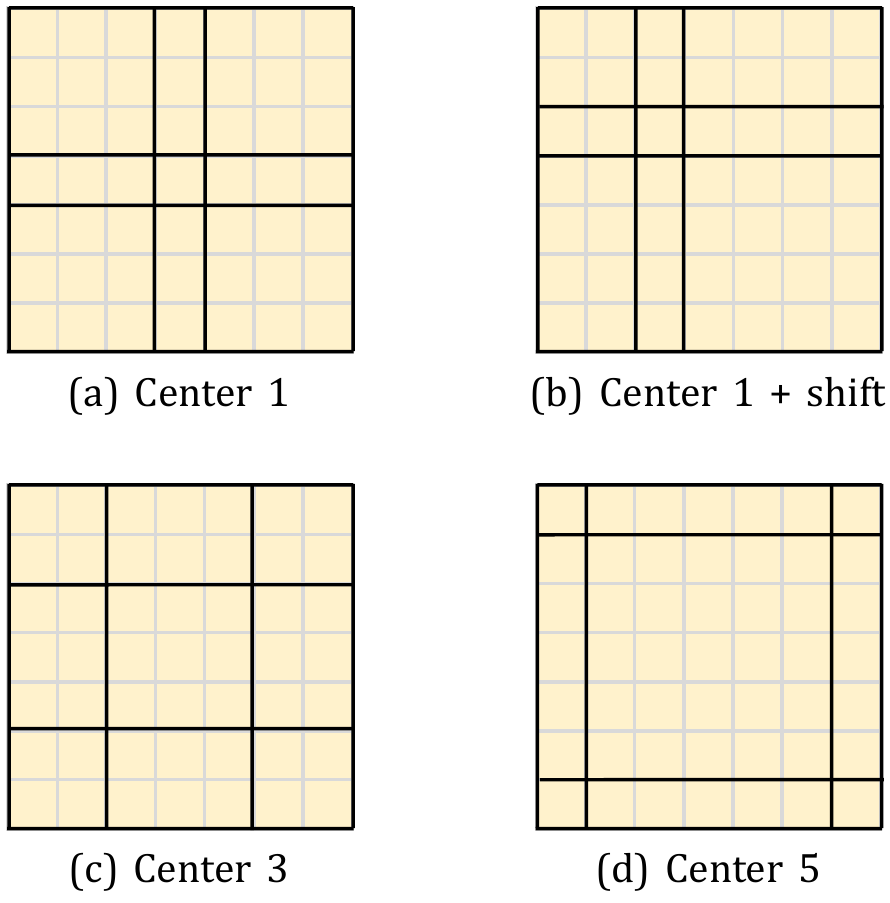}
   \caption{The center group manners in Tab.~~\textcolor{red}{4} in the paper. We study the center sizes for group splitting and center shifting.}
   \label{fig:kernel-groups}
\end{center}
\end{figure}

In this supplementary material, we first introduce implementation details in Sec.~\ref{sec:implementation-details}. It includes data processing, training settings, and network architectures. After that, in Sec.~\ref{sec:experiments-results}, we introduce additional experimental results on the nuScenes dataset. We also provide additional visualizations on effective receptive fields~(ERFs) in Sec.~\ref{sec:visualizations} and the illustration manner. Note that the rank of LargeKernel3D on nuScenes {\em test} is reported at the paper submission time. Methods that released afterwards are not counted. We visualize group manners of Tab.~~\textcolor{red}{4} in the paper in Fig.~S~-~\ref{fig:kernel-groups}.

\section{More Implementation Details}
\label{sec:implementation-details}
\subsection{Data Processing}
\noindent
\textbf{ScanNetv2}
We convert point clouds into voxels as input data for the ScanNetv2 dataset. The voxelization sizes are all 0.02m for all {\em X}, {\em Y}, {\em Z} axes. In terms of data augmentations, we exactly follow our baseline method, MinkowskiNet~\cite{minkowskinet}. Specially, input data is randomly dropped out with a ratio of 0.2. For spatial augmentations, we also conduct random horizontal flipping. For intensity augmentations, we conduct auto-contrast, color translation, and jittering.
\vspace{0.5em}
\noindent
\textbf{nuScenes}
We convert point clouds into voxels as input data.
We clip point clouds into [-54m, 54m] for both {\em X} and {\em Y} axes, and [-5m, 3m] for the {\em Z} axis, on nuScenes~\cite{nuscenes}. The voxel size is set as (0.075m, 0.075m, 0.2m). Data augmentations include random flipping, global scaling, global rotation, GT sampling~\cite{second}, and an additional translation on the nuScenes~\cite{nuscenes} dataset. Random flipping is performed in {\em X} and {\em Y} axes. Rotation angle is sampled in [-45$^{\mathrm{o}}$, 45$^{\mathrm{o}}$]. Global scaling is conducted in the [0.9, 1.1] ratio. Translation noise is conducted on all three axes from the ratio [0, 0.5]. GT sampling is also conducted on the nuScenes.
\begin{table*}[t]
\begin{center}
\caption{{Comparison with other methods and the ground-truth sampling fading~(GT-S Fading) trick on the nuScenes {\em val} split.}}
\begin{tabular}{|l|cc|cccccccccc|}
\hline
Method        & NDS    & mAP  & Car                  & Truck                & Bus                  & Trailer              & C.V.                   & Ped                  & Mot                  & Byc                  & T.C.                   & Bar                   \\ \hline
CenterPoint~\cite{centerpoint}   & 66.4 & 59.0 & 85.6 & 57.2 & 71.2 & 37.3 & 16.2 & 85.1 & 58.4 & 41.0 & 69.2 & 68.2 \\ 
TransFusion~\cite{transfusion}     & 66.8  & 60.0 & 85.8 & 57.6 & 71.6 & 37.3 & 19.3 & 86.7 & 57.2 & 42.3 & 71.0 & 69.7 \\ 
Focals Conv~\cite{focal-sparse-conv} & 67.2 & 60.2 & 85.7 & 58.4 & 71.5 & 37.8 & 19.0 & 85.5  & 58.5 & 45.6 & 70.4 & 69.2 \\ \hline
LargeKernel3D & \textbf{67.5} & \textbf{60.3} & 85.2 & 58.3 & 71.6 & 37.9 & 19.8 & 85.4 & 60.8 & 44.3 & 70.7 & 68.6 \\  
+ GT-S Fading & \textbf{69.1} & \textbf{63.9} &  85.1 & 60.1 & 72.6 & 41.4 & 24.3 & 85.6 & 70.8 & 59.2 & 72.3 & 67.7 \\  \hline
\end{tabular}
\label{tab:nuscenes-val}
\end{center}
\end{table*}
\begin{figure*}[t]
\begin{center}
   \includegraphics[width=\linewidth]{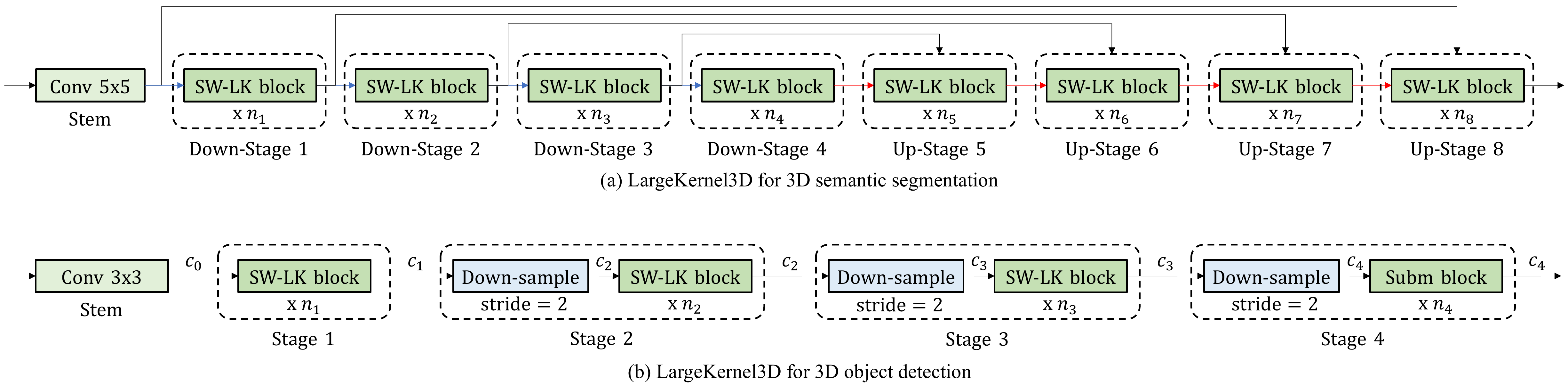}
      \caption{Architectures of LargeKernel3D for 3D semantic segmentation and object detection.}\label{fig:sw-lknet}
      \includegraphics[width=\linewidth]{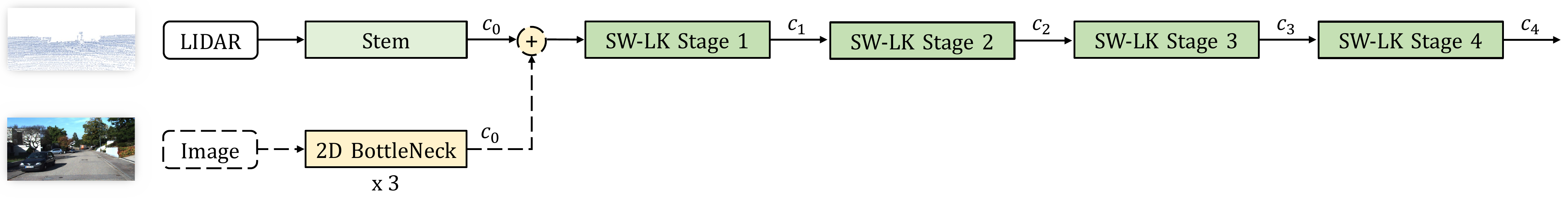}
   \caption{Architectures of LargeKernel3D-F with image fusion for 3D object detection.}\label{fig:imagefusion}
\end{center}
\end{figure*}

\begin{figure*}
\begin{center}
   \includegraphics[width=\linewidth]{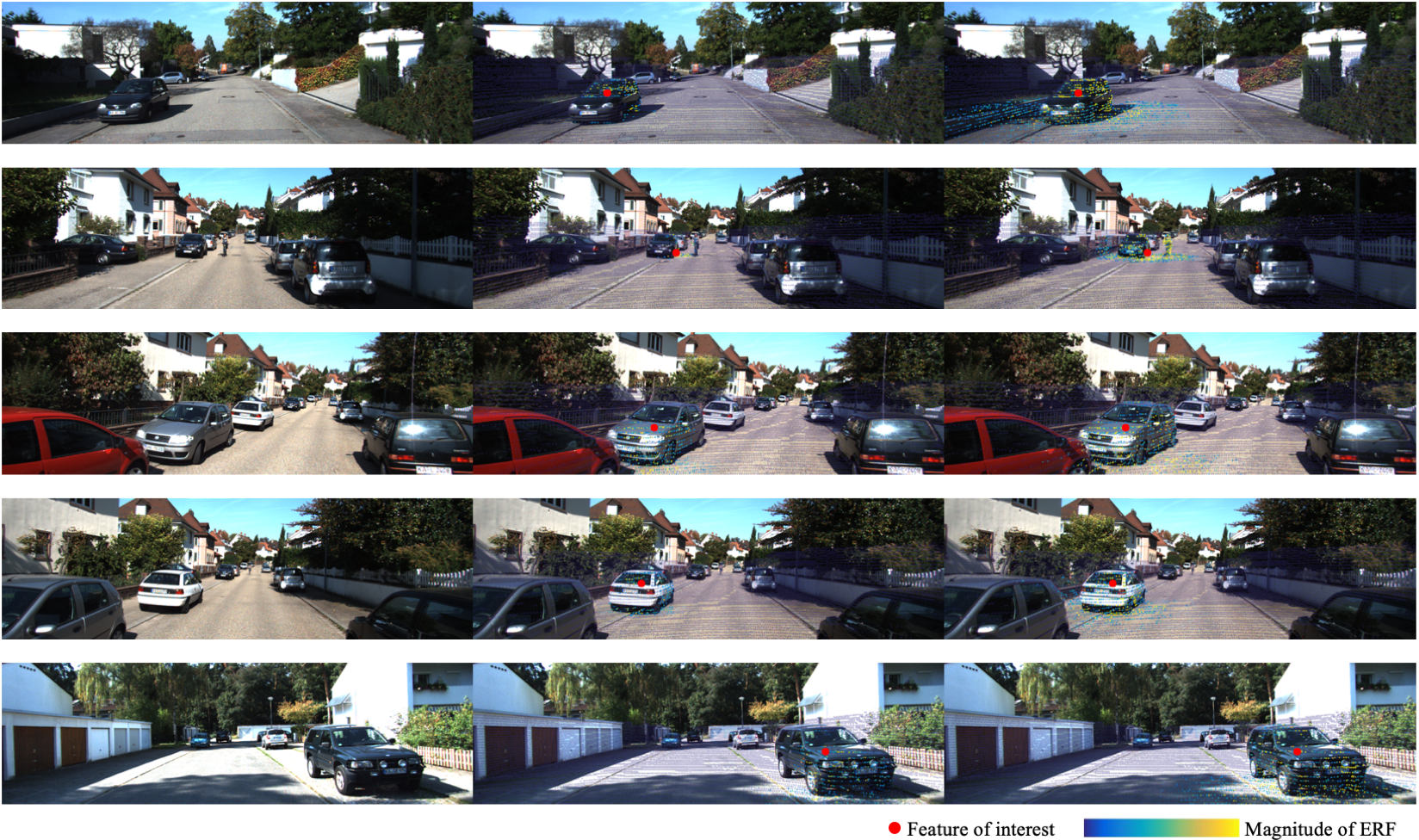}
   \caption{Additional illustrations on effective receptive fields. Left - original images, mid - plain 3D CNNs, right - LargeKernel3D. It is best viewed in color and by zooming in.}
   \label{fig:effective-receptive-fields-more}
\end{center}
\end{figure*}

\vspace{0.5em}
\noindent
\textbf{Waymo}
Point clouds is clipped into [-75.2m, 75.2m] {\em X} and {\em Y} axis, and [-2m, 4m] for {\em Z} axis, on Waymo~\cite{waymo} for ranges. The input voxel size is set as (0.1m, 0.1m, 0.15m). 
The data augmentations include random flipping, global scaling, global rotation, and ground-truth (GT) sampling~\cite{second} for the Waymo dataset. Random flipping is applied along {\em X} and {\em Y} axes. Global scaling is sampled from the [0.95, 1.05] ratio. Global rotation is performed around the {\em Z} axis. Rotation angle is sampled from [-45$^{\mathrm{o}}$, 45$^{\mathrm{o}}$]. Ground-truth sampling copies objects from other training data, and pastes them onto the current scene. It enriches data variance during training.  These settings follow baseline methods~\cite{centerpoint,pvrcnn}.

\subsection{Training Settings}
\noindent
\textbf{ScanNetv2}
For models trained on the ScanNetv2 dataset, we train networks for 600 epochs with batch size 16. The learning rate is initialized as 0.1 and decays with a poly scheduler. We adopt SGD optimizer. The momentum is set as 0.9. Hyper-parameters directly follow our baseline~\cite{minkowskinet}.

\vspace{0.5em}
\noindent
\textbf{nuScenes}
We train CenterPoint~\cite{centerpoint} on the nuScenes datasets for 20 epochs with batch size 32. This network is trained by Adam. The learning rate is set as 1e-3 and decays in the cosine annealing strategy to 1e-4. The weight decay is set as 0.01. The gradient norms are clipped by 35.

\vspace{0.5em}
\noindent
\textbf{Waymo}
We train the network for 30 epochs and batch size 16 on Waymo. The learning rate is initialized as 0.003. Gradient norms are clipped by 10. We adopt the Adam optimizer, with weight decay 0.01 and momentum 0.9. These settings follow the CenterPoint~\cite{centerpoint} baseline.

\subsection{Network Architecture Settings}
\noindent
\textbf{3D Semantic Segmentation}
We use MinkowskiNet-34~\cite{minkowskinet} as the baseline, for the ScanNetv2 dataset in the paper. In MinkowskiNet-34, we set the channel numbers as \{32, 64, 128, 256, 256, 128, 96, 96\}. The block numbers, \{$n_1$, $n_2$, $n_3$, $n_4$, $n_5$, $n_6$, $n_7$, $n_8$\}, are \{2, 3, 4, 6, 2, 2, 2, 2\}. The meanings of these notations are shown in Fig.~S~-~\ref{fig:sw-lknet}.
LargeKernel3D directly follows MinkowskiNet-34 for these settings. They substitute the plain sparse convolutional blocks to the proposed SW-LK Conv with spatial size 7$^3$ and groups 3$^3$. The tiny version, LargeKernel3D-T, has half channel numbers of the original in the last two stages.

\vspace{0.5em}
\noindent
\textbf{3D Object Detection}
The backbone network of CenterPoint~\cite{centerpoint} has channels \{$c_0$, $c_1$, $c_2$, $c_3$, $c_4$\}, equal to \{16, 16, 32, 64, 128\}. The block numbers in these stages, \{$n_1$, $n_2$, $n_3$, $n_4$\}, are \{2, 2, 2, 2\}. Each block contains two convolutional layers, with a residual connection, except the stem. LargeKernel3D also substitutes the plain blocks for SW-LK blocks for stages 1, 2, 3. Because the last stage has heavy channel numbers and its receptive field is already sufficient.

We also present the multi-modal network with our large kernel backbone, {\em i.e.}, LargeKernel3D-F. As shown in Fig.~S~-~\ref{fig:imagefusion}, we conduct a direct voxel-wise summation between LIDAR and RGB features. The RGB branch only contains a conv-bn-relu-pooling stem and 3 residual bottlenecks~\cite{resnet}. We intentionally make the RGB branch lightweight to fully demonstrate the capacity of our large-kernel LIDAR backbone. These settings follow~\cite{focal-sparse-conv}.

\section{Additional Experimental Results}
\label{sec:experiments-results}
We present further improvements on the nuScenes dataset by additional techniques in Tab.~\ref{tab:nuscenes-val}. These techniques are removing gt-sampling in the last 5 training epochs~(GT-S Fading). This trick has been used by some previous state-of-the-art methods~\cite{pointaugmenting, focal-sparse-conv} for performance boosting. As shown in Tab.~\ref{tab:nuscenes-val}, LargeKernel3D achieves 63.9\% mAP on the {\em val} split. This technique is included for test submission. For the multi-modal LargeKernel3D-F, it has the potential to achieve better performance if equipped with more advanced and heavier fusion methods~\cite{bevfusion-liu,bevfusion-liang,deepinteraction}. We would like to try these extensions in future work.

\section{Visualizations}
\label{sec:visualizations}
We provide additional visual comparisons between the plain 3D network and our LargeKernel3D in Fig.~S~-~\ref{fig:effective-receptive-fields-more}. It shares the same setting as the Fig.~\textcolor{red}{2} in the paper. In each group, the left one is the original image, the middle one is the ERFs of plain 3D CNNs, and the right one is the ERFs of our LargeKernel3D. We follow the definition of ERFs~\cite{effective-receptive-fields}. We calculate the gradient of every input voxel data regarding the feature of interest. It illustrates the intensity of feature changes as the input value changes. We normalize the gradient norms to [0, 1] and project them onto the image plane via calibration matrices.

\end{document}